\documentclass[letterpaper, 10 pt, conference]{ieeeconf}  
\IEEEoverridecommandlockouts
\usepackage{graphicx}
\usepackage{caption}
\graphicspath{
    {graphics/}
}

\usepackage{amsmath,amssymb,enumerate}

\usepackage{amsthm}
\usepackage{setspace}
\usepackage{booktabs}
\usepackage[usenames,dvipsnames,svgnames,table]{xcolor}
\usepackage[colorlinks=true,pdfpagemode=UseNone,citecolor=black,linkcolor=black,urlcolor=BrickRed]{hyperref}
\usepackage{mathtools}
\usepackage{algorithm, algorithmicx, algpseudocode}
\usepackage{blindtext}
\usepackage{gensymb}
\usepackage{xparse}
\usepackage{lipsum}
\usepackage{graphicx}
\usepackage{soul} 
\usepackage{mathrsfs}
\usepackage[mathscr]{euscript}
\usepackage{times}
\usepackage{cite} % to group references
\usepackage{multicol}
\usepackage[caption=false,font=footnotesize]{subfig}
\usepackage{amsfonts}
\usepackage{fixltx2e}
\usepackage{cleveref}
\usepackage[utf8]{inputenc}
\usepackage[T1]{fontenc}
\usepackage{textcomp}
\usepackage{balance}

\newtheorem{theorem}{Theorem}

\newtheorem{proposition}[theorem]{Proposition}

\newtheorem{remark}{Remark}

\def\realnumbers{\mathbb{R}}

\DeclareMathOperator*{\minimize}{minimize}

\renewcommand{\thefigure}{\arabic{figure}}

\definecolor{note}{rgb}{0.1,0.1,1}
\definecolor{rephase}{rgb}{0.15,0.7,0.15}
\definecolor{bag}{rgb}{0.6,0.6,0.2}

\newcommand{\Covariance}[1]{\mathbf{Cov}[#1]} % Covariance

\newcommand{\Hz}{\mathop{\mathrm{Hz}}}

\newcommand{\Ccal}{\mathcal{C}}

\newcommand{\Fcal}{\mathcal{F}}
\newcommand{\Hcal}{\mathcal{H}}
\newcommand{\Ical}{\mathcal{I}}
\newcommand{\Kcal}{\mathcal{K}}
\newcommand{\Lcal}{\mathcal{L}}

\newcommand{\Scal}{\mathcal{S}}
\newcommand{\Tcal}{\mathcal{T}}
\newcommand{\Xcal}{\mathcal{X}}

\newcommand{\Zcal}{\mathcal{Z}}

\newcommand{\Exp}{\mathrm{Exp}}
\newcommand{\Log}{\mathrm{Log}}

\newcommand{\transpose}{\mathsf{T}}
\newcommand{\SO}{\mathrm{SO}}
\newcommand{\SE}{\mathrm{SE}}
\newcommand{\GL}{\mathrm{GL}}

\DeclareDocumentCommand{\vectorToSkew}{ O{} }{\left(#1\right)_\times}

\DeclareDocumentCommand{\vector}{ O{} }{\mathrm{vec}(#1)}
\DeclareDocumentCommand{\zeros}{ O{} }{\textbf{0}_{#1}}
\DeclareDocumentCommand{\X}{ O{} O{} }{\textbf{X}_{#1}^{#2}}
\DeclareDocumentCommand{\XE}{ O{} O{} }{\hat{\textbf{X}}_{#1}^{#2}}
\DeclareDocumentCommand{\L}{ O{} O{} }{\textbf{L}_{#1}^{#2}}
\DeclareDocumentCommand{\S}{ O{} O{} }{\textbf{S}_{#1}^{#2}}
\DeclareDocumentCommand{\P}{ O{} O{} }{\textbf{P}_{#1}^{#2}}
\DeclareDocumentCommand{\B}{ O{} O{} }{\textbf{B}_{#1}^{#2}}
\DeclareDocumentCommand{\Q}{ O{} O{} }{\textbf{Q}_{#1}^{#2}}
\DeclareDocumentCommand{\H}{ O{} O{} }{\textbf{H}_{#1}^{#2}}
\DeclareDocumentCommand{\J}{ O{} O{} }{\textbf{J}_{#1}^{#2}}
\DeclareDocumentCommand{\R}{ O{} O{} }{\textbf{R}_{#1}^{#2}}
\DeclareDocumentCommand{\RE}{ O{} O{} }{\hat{\textbf{R}}_{#1}^{#2}}
\DeclareDocumentCommand{\Q}{ O{} O{} }{\textbf{Q}_{#1}^{#2}}
\DeclareDocumentCommand{\T}{ O{} O{} }{\textbf{T}_{#1}^{#2}}
\DeclareDocumentCommand{\L}{ O{} O{} }{\textbf{L}_{#1}^{#2}}
\DeclareDocumentCommand{\K}{ O{} O{} }{\textbf{K}_{#1}^{#2}}
\DeclareDocumentCommand{\V}{ O{} O{} }{\textbf{V}_{#1}^{#2}}
\DeclareDocumentCommand{\N}{ O{} O{} }{\textbf{N}_{#1}^{#2}}
\DeclareDocumentCommand{\Y}{ O{} O{} }{\textbf{Y}_{#1}^{#2}}
\DeclareDocumentCommand{\F}{ O{} O{} }{\textbf{F}_{#1}^{#2}}
\DeclareDocumentCommand{\G}{ O{} O{} }{\textbf{G}_{#1}^{#2}}
\DeclareDocumentCommand{\A}{ O{} O{} }{\textbf{A}_{#1}^{#2}}
\DeclareDocumentCommand{\TH}{ O{} O{} }{\boldsymbol{\Theta}_{#1}^{#2}}
\DeclareDocumentCommand{\x}{ O{} O{} }{\textbf{x}_{#1}^{#2}}
\DeclareDocumentCommand{\e}{ O{} O{} }{\textbf{e}_{#1}^{#2}}
\DeclareDocumentCommand{\c}{ O{} O{} }{\textbf{c}_{#1}^{#2}}
\DeclareDocumentCommand{\C}{ O{} O{} }{\textbf{C}_{#1}^{#2}}
\DeclareDocumentCommand{\CM}{ O{} O{} }{\tilde{\textbf{C}}_{#1}^{#2}}
\DeclareDocumentCommand{\RM}{ O{} O{} }{\tilde{\textbf{R}}_{#1}^{#2}}
\DeclareDocumentCommand{\I}{ O{} O{} }{\textbf{I}_{#1}^{#2}}
\DeclareDocumentCommand{\O}{ O{} O{} }{\textbf{O}_{#1}^{#2}}
\DeclareDocumentCommand{\r}{ O{} O{} }{\textbf{r}_{#1}^{#2}}
\DeclareDocumentCommand{\t}{ O{} O{} }{\textbf{t}_{#1}^{#2}}
\DeclareDocumentCommand{\u}{ O{} O{} }{\textbf{u}_{#1}^{#2}}
\DeclareDocumentCommand{\d}{ O{} O{} }{\textbf{d}_{#1}^{#2}}
\DeclareDocumentCommand{\dE}{ O{} O{} }{\hat{\textbf{d}}_{#1}^{#2}}
\DeclareDocumentCommand{\b}{ O{} O{} }{\textbf{b}_{#1}^{#2}}
\DeclareDocumentCommand{\a}{ O{} O{} }{\textbf{a}_{#1}^{#2}}
\DeclareDocumentCommand{\g}{ O{} O{} }{\textbf{g}_{#1}^{#2}}
\DeclareDocumentCommand{\dM}{ O{} O{} }{\tilde{\textbf{d}}_{#1}^{#2}}
\DeclareDocumentCommand{\params}{ O{} O{} }{\boldsymbol{\theta}_{#1}^{#2}}
\DeclareDocumentCommand{\paramsE}{ O{} O{} }{\hat{\boldsymbol{\theta}}_{#1}^{#2}}
\DeclareDocumentCommand{\paramError}{ O{} O{} }{\boldsymbol{\zeta}_{#1}^{#2}}
\DeclareDocumentCommand{\gyroscopeBias}{ O{} }{\textbf{b}_{#1}^{g}}
\DeclareDocumentCommand{\gyroscopeBiasE}{ O{} }{\hat{\textbf{b}}_{#1}^{g}}
\DeclareDocumentCommand{\accelerometerBias}{ O{} }{\textbf{b}_{#1}^{a}}
\DeclareDocumentCommand{\accelerometerBiasE}{ O{} }{\hat{\textbf{b}}_{#1}^{a}}
\DeclareDocumentCommand{\position}{ O{} O{} }{{}_\text{#2}\textbf{p}_{\text{#1}}}
\DeclareDocumentCommand{\positionDot}{ O{} O{} }{{}_\text{#2}\dot{\textbf{p}}_{\text{#1}}}
\DeclareDocumentCommand{\linearVelocity}{ O{} O{} }{{}_\text{#2}\textbf{v}_{\text{#1}}}
\DeclareDocumentCommand{\linearVelocityM}{ O{} O{} }{{}_\text{#2}\tilde{\textbf{v}}_{\text{#1}}}
\DeclareDocumentCommand{\orientation}{ O{} }{\textbf{R}_{\text{#1}}}
\DeclareDocumentCommand{\orientationDot}{ O{} }{\dot{\textbf{R}}_{\text{#1}}}
\DeclareDocumentCommand{\homogeneous}{ O{} O{} }{{}_\text{#2}\boldsymbol{\H}_{\text{#1}}}
\DeclareDocumentCommand{\homogeneousDot}{ O{} }{\dot{\textbf{H}}_{\text{#1}}}
\DeclareDocumentCommand{\Jacobian}{ O{} O{} }{{}_\text{#2}\boldsymbol{\J}_{\text{#1}}}
\DeclareDocumentCommand{\inputs}{ O{} O{} }{{}_\text{#2}\boldsymbol{\u}_{\text{#1}}}
\DeclareDocumentCommand{\inputsM}{ O{} O{} }{{}_\text{#2}\tilde{\boldsymbol{\u}}_{\text{#1}}}
\DeclareDocumentCommand{\angularVelocity}{ O{} O{} }{{}_\text{#2}\boldsymbol{\omega}_{\text{#1}}}
\DeclareDocumentCommand{\angularVelocityM}{ O{} O{} }{{}_\text{#2}\tilde{\boldsymbol{\omega}}_{\text{#1}}}
\DeclareDocumentCommand{\acceleration}{ O{} O{} }{{}_\text{#2}\textbf{a}_{\text{#1}}}
\DeclareDocumentCommand{\accelerationM}{ O{} O{} }{{}_\text{#2}\tilde{\textbf{a}}_{\text{#1}}}
\DeclareDocumentCommand{\p}{ O{} O{} }{\textbf{p}_{#1}^{#2}}
\DeclareDocumentCommand{\pE}{ O{} O{} }{\hat{\textbf{p}}_{#1}^{#2}}
\DeclareDocumentCommand{\pM}{ O{} O{} }{\tilde{\textbf{p}}_{#1}^{#2}}
\DeclareDocumentCommand{\v}{ O{} O{} }{\textbf{v}_{#1}^{#2}}
\DeclareDocumentCommand{\vE}{ O{} O{} }{\hat{\textbf{v}}_{#1}^{#2}}
\DeclareDocumentCommand{\w}{ O{} O{} }{\boldsymbol{\omega}_{#1}^{#2}}
\DeclareDocumentCommand{\wM}{ O{} O{} }{\tilde{\boldsymbol{\omega}}_{#1}^{#2}}
\DeclareDocumentCommand{\noise}{ O{} O{} }{\textbf{w}_{#1}^{#2}}
\DeclareDocumentCommand{\FK}{ O{} }{\;\textit{\textbf{h}}_{#1}}
\DeclareDocumentCommand{\FKswitch}{ O{} }{\;\textit{\textbf{g}}_{#1}}
\DeclareDocumentCommand{\angleTheta}{ O{} O{} }{\boldsymbol{\theta}_{#1}^{#2}}
\DeclareDocumentCommand{\anglePhi}{ O{} O{} }{\boldsymbol{\phi}_{#1}^{#2}}
\DeclareDocumentCommand{\encoders}{ O{} O{} }{\boldsymbol{\alpha}_{#1}^{#2}}
\DeclareDocumentCommand{\encodersM}{ O{} O{} }{\tilde{\boldsymbol{\alpha}}_{#1}^{#2}}
\DeclareDocumentCommand{\offsets}{ O{} O{} }{\boldsymbol{\epsilon}_{#1}^{#2}}
\DeclareDocumentCommand{\Cov}{ O{} O{} }{\boldsymbol{\Sigma}_{#1}^{#2}}
\DeclareDocumentCommand{\Axis}{ O{} O{} }{\textnormal{Axis}_{#1}^{#2}}
\DeclareDocumentCommand{\using}{ O{} O{} }{\stackrel{\mathmakebox[\widthof{=}]{\text{eq.} (#1)}}{#2} \enspace}
\DeclareDocumentCommand{\Adjoint}{ O{} }{\mathrm{Ad}_{#1}}
\DeclareDocumentCommand{\vectorToAlgebra}{ O{} }{\mathscr{L}_\mathfrak{g}\left(#1\right)}
\DeclareDocumentCommand{\groupError}{ O{} O{} }{\boldsymbol{\eta}_{#1}^{#2}}
\DeclareDocumentCommand{\twist}{ O{} O{} }{\boldsymbol{\xi}_{#1}^{#2}}
\DeclareDocumentCommand{\twistM}{ O{} O{} }{\tilde{\boldsymbol{\xi}}_{#1}^{#2}}

\DeclareDocumentCommand{\JFK}{ O{} }{\textbf{J}^{\hspace{-0.75mm}\FK}_{#1}}

\DeclareDocumentCommand{\JFKswitch}{ O{} }{\textbf{J}^{\hspace{-0.75mm}\FKswitch}_{#1}}

\newcommand{\squeezeup}{\vspace{-3mm}}

\makeatletter
\newcommand{\mathleft}{\@fleqntrue\@mathmargin0pt}
\newcommand{\mathcenter}{\@fleqnfalse}
\makeatother

\title{\LARGE \bf 
Hybrid Contact Preintegration for Visual-Inertial-Contact State Estimation Using Factor Graphs
}

\author{Ross Hartley, Maani Ghaffari Jadidi, Lu Gan, Jiunn-Kai Huang, Jessy W. Grizzle, and Ryan M. Eustice
\thanks{The authors are with the College of Engineering, University of Michigan, Ann Arbor, MI 48109 USA {\tt\small \{{rosshart, maanigj, ganlu, bjhuang, grizzle, eustice\}}@umich.edu}.}
}

\begin{document}
\maketitle
\begin{abstract} 
The factor graph framework is a convenient modeling technique for robotic state estimation where states are represented as nodes, and measurements are modeled as factors. When designing a sensor fusion framework for legged robots, one often has access to visual, inertial, joint encoder, and contact sensors. While visual-inertial odometry has been studied extensively in this framework, the addition of a preintegrated contact factor for legged robots has been only recently proposed. This allowed for integration of encoder and contact measurements into existing factor graphs, however, new nodes had to be added to the graph every time contact was made or broken. In this work, to cope with the problem of switching contact frames, we propose a hybrid contact preintegration theory that allows contact information to be integrated through an arbitrary number of contact switches. The proposed hybrid modeling approach reduces the number of required variables in the nonlinear optimization problem by only requiring new states to be added alongside camera or selected keyframes. This method is evaluated using real experimental data collected from a Cassie-series robot where the trajectory of the robot produced by a motion capture system is used as a proxy for ground truth. The evaluation shows that inclusion of the proposed preintegrated hybrid contact factor alongside visual-inertial navigation systems improves estimation accuracy as well as robustness to vision failure, while its generalization makes it more accessible for legged platforms.

%Supplementary file available at:\href{https://www.dropbox.com/sh/urik6o4ksxtsjfp/AADqCvcweaqcjKsfhpXA3Ptxa?dl=0}{temporary submission link}
\end{abstract}

\thispagestyle{empty}
\pagestyle{empty}

%%%%%%%%%%%%%%%%%%%%%%%%%%
\section{Introduction and Related Work}
\label{sec:intro}

Long-term state estimation and mapping for legged robots requires a flexible sensor fusion framework that allows for reducing drift and correcting past estimates as the robot perceives new information. During long-term missions, odometry systems can drift substantially since the absolute position and yaw (rotation about gravity) are unobservable~\cite{bloesch2013state,Hartley-RSS-18}, leading to an unbounded growth in the covariance of the estimate and an undesirable expansion of the search space for data association tasks. The factor graph smoothing framework~\cite{kaess2012isam2,dellaert2012factor,carlone2014eliminating,forster2016manifold} offers suitable machineries for building such systems in which real-time performance is achieved by exploiting the sparse structure of the Simultaneous Localization and Mapping (SLAM) problem~\cite{thrun2004simultaneous,eustice2006exactly}. In addition, the incorporation of loop-closures~\cite{eustice2006exactly} into the graph, upon availability, is convenient. 

Legged robot perception often involves fusing kinematic odometry, Inertial Measurement Unit (IMU) data, and visual/depth measurements to infer the robot's trajectory, controller inputs such as velocity, and calibration parameters~\cite{rotella2014state,benallegue2015estimation,eljaik2015multimodal,kuindersma2016optimization}.
The challenge in such perception problems is the rigorous real-time performance requirements in legged robots arising from their direct and switching contact with the environment~\cite{bloesch2013state,fankhauser2014robot,bloesch2017state,fallon2014drift,nobiliheterogeneous}. Furthermore, kinematic leg odometry involves estimating relative transformations and velocity using kinematic and contact information, which can be noisy due to the encoder noise and foot slip~\cite{roston1991dead,Hartley-RSS-18}.

\begin{figure}[t]
\vspace{.25cm}
  \centering 
  \includegraphics[width=0.9\columnwidth]{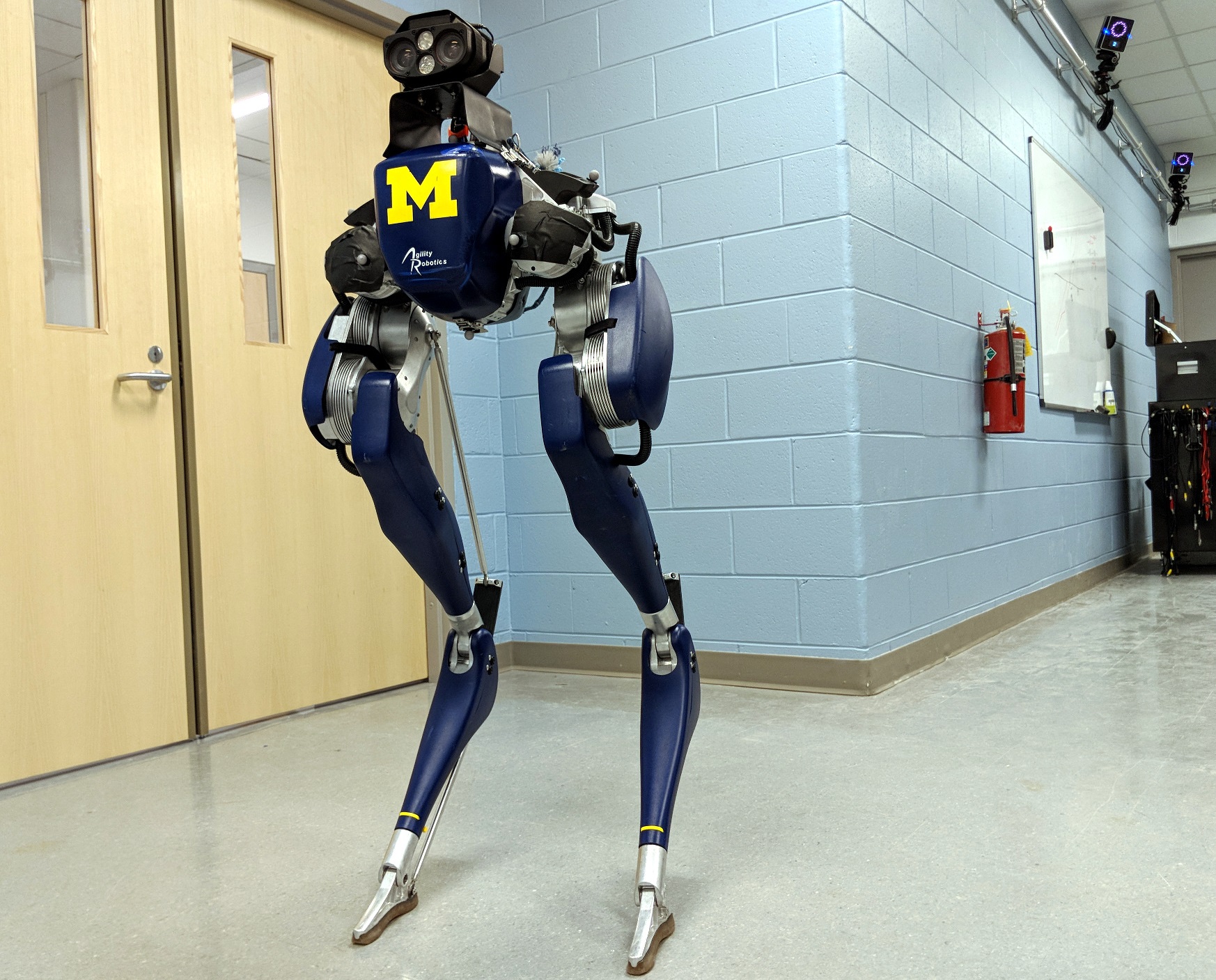}
  \caption{Experiments were conducted on a Cassie-series robot designed by Agility Robotics in an indoor laboratory environment. The motion capture system is used to record the robot trajectory as a proxy for ground truth data. The Cassie-series robot has 20 degrees of freedom, 10 actuators, joint encoders, an IMU, and mounted with a Multisense S7 stereo camera.}
  \label{fig:first_fig}
  \squeezeup\squeezeup
\end{figure}

In our previous work~\cite{rhartley-2018a}, we made progress towards a long-term, legged robot state estimator by developing two novel factors that integrate the use of forward kinematics and the notion of contact between the robot and the environment into the factor graph framework. While these novel factors improved the estimator's performance, new challenges emerged due to frequent switching contacts as the robot navigates through an environment. In particular, a new node needs to be added to the graph every time contact was made or broken. This leads to an larger number of optimization variables, which increases the complexity of the graph and ultimately slows down the estimator. This problem is only exacerbated as the number of contact points increases (quadraped or hexapod robots).

In this paper, we generalize the idea of the preintegrated contact factor in~\cite{rhartley-2018a} to a \emph{hybrid preintegrated contact factor} that alleviates the effect of frequently changing contacts with the environment. We develop a novel method for preintegrating contact information though an arbitrary number of contact switches. The proposed hybrid modeling approach reduces the number of required variables in the nonlinear optimization problem by only requiring new states to be added alongside camera or selected keyframes. The present work has the following contributions:

\begin{enumerate}[i.]
\item A generic forward kinematic factor modeled in $\SE(3)$ using the manipulator Jacobian that supports both prismatic and revolute joints; the factor incorporates noisy encoder measurements to estimate an end-effector pose at any time-step.
\item A hybrid preintegrated contact factor modeled in $\SE(3)$ that allows for an arbitrary number of contact switches between camera or selected keyframes.
\item Real-time implementation and experimental evaluation of the derived factors on a Cassie-series biped robot in an indoor laboratory environment where ground truth data was collected using a motion capture system.
\end{enumerate}

The remainder of this paper is organized as follows. Section~\ref{sec:preliminaries} provides the mathematical background and preliminaries. We formulate the problem using the factor graph approach in Section~\ref{sec:problem}. Section~\ref{sec:fk} explains the forward kinematic modeling and derives the forward kinematic factor. The proposed hybrid rigid contact model and hybrid contact preintegration are derived in Section~\ref{sec:contact_preintegration}. Experimental evaluations of the proposed methods on a 3D biped robot (shown in Fig. \ref{fig:first_fig}) are presented in Section~\ref{sec:results}. Finally, Section~\ref{sec:conclusion} concludes the paper and provides future work suggestions.

\section{Mathematical Background and Preliminaries}
\label{sec:preliminaries}

We first review the Lie group theory corresponding to the rotation and motion groups~\cite{murray1994mathematical,chirikjian2011stochastic}. Afterwards, we discuss the optimization technique on matrix Lie groups~\cite{absil2009optimization} and the choice of retraction for deriving forward kinematic and hybrid preintegrated contact factors.

Matrices are capitalized in bold, such as in $\X$, and vectors are column-wise in lower case bold type, such as in $\x$. We denote $\lVert \boldsymbol \e \rVert_{\Cov}^2 \triangleq \e[][\transpose] \Cov[][-1] \e$. The $n$-by-$n$ identity matrix and the $n$-by-$m$ matrix of zeros are denoted by $\I[n]$ and $\zeros[n \times m]$ respectively. The vector constructed by stacking $x_i$, \mbox{$\forall \ i \in \{1,\dots,n\}$} is denoted by $\mathrm{vec}(x_1,\dots,x_n)$. The covariance of a random vector is denoted by $\Covariance{\cdot}$. Finally, we denote the base (IMU) frame of the robot by $\textnormal{B}$, the world frame by $\textnormal{W}$, and contact frame by $\textnormal{C}$.

\subsection{Matrix Lie Group of Rotation and Motion in $\mathbb{R}^3$}

The general linear group of degree $n$, denoted by \mbox{\small{$\GL_n(\mathbb{R})$}}, is the set of all $n\times n$ nonsingular real matrices, where the group binary operation is the ordinary matrix multiplication. The three-dimensional (3D) special orthogonal group, denoted by {$\SO(3) = \{\R\in \GL_3(\realnumbers)\ |\ \R \R[][\transpose] = \mathbf{I}_3, \operatorname{det} \R = +1\}$ is the rotation group on $\mathbb{R}^3$. The 3D special Euclidean group, denoted by
\begin{equation}
%\resizebox{\columnwidth}{!}{$
\small
\nonumber \SE(3) = \{ \homogeneous = \begin{bmatrix} \R & \p \\ \zeros[1 \times 3] & 1 \end{bmatrix} \in \GL_4(\realnumbers)\ |\ \R \in \SO(3), \p \in \realnumbers^3 \}%$}
\end{equation}
is the group of rigid transformations on $\realnumbers^3$. The Lie algebra (tangent space at the identity together with Lie bracket) of $\SO(3)$, denoted by $\mathfrak{so}(3)$, is the set of $3\times 3$ skew-symmetric matrices such that for any $\angularVelocity \triangleq \mathrm{vec}(\omega_1, \omega_2, \omega_3) \in \realnumbers^3$:
\begin{equation}
\small
 \nonumber \boldsymbol \omega^\wedge \triangleq 
\begin{bmatrix}
0 & -\omega_3 & \omega_2 \\
\omega_3 & 0 & - \omega_1 \\
-\omega_2 & \omega_1 & 0 \\
\end{bmatrix},
\end{equation}
and ${({\angularVelocity}^\wedge)}^{\vee} = \angularVelocity$. The Lie algebra of $\SE(3)$, denoted by $\mathfrak{se}(3)$, can be identified by $4\times 4$ matrices such that for any $\angularVelocity, \v \in \realnumbers^3$ and the twist is defined as $\twist \triangleq \mathrm{vec}(\angularVelocity, \v) \in \mathbb{R}^6$:
\begin{equation}
\label{eq:twist}
 \twist[][\wedge] \triangleq 
\begin{bmatrix}
\angularVelocity^\wedge & \v \\
\zeros[1 \times 3] & 0 \\
\end{bmatrix}
\end{equation}
where the wedge operator ($^\wedge$) for twist is overloaded.

The exponential map $\exp:\mathfrak{se}(3) \to \SE(3)$ can be used to map a member of $\mathfrak{se}(3)$ around a neighborhood of zero to a member of $\SE(3)$ around a neighborhood of the identity. The logarithm map is the inverse, i.e., $\log:\SE(3) \to \mathfrak{se}(3)$, and $\exp(\log(\homogeneous)) = \homogeneous$, $\homogeneous \in \SE(3)$. Now we can define the difference between a transformation $\homogeneous \in \SE(3)$ and its estimate with a small perturbation $\tilde{\homogeneous} \in \SE(3)$ as~\cite{chirikjian2011stochastic,barfoot2014tro}:
\begin{equation}
 \nonumber \boldsymbol \epsilon^{\wedge} = \log(\homogeneous^{-1} \tilde{\homogeneous}),
\end{equation}
where $\boldsymbol \epsilon^{\wedge} \in \mathfrak{se}(3)$. To define the norm and covariance of the error term, we exploit the fact that $\mathfrak{se}(3)$ is isomorphic to $\mathbb{R}^6$, i.e., $\boldsymbol \epsilon^{\wedge} \mapsto \boldsymbol \epsilon \in \mathbb{R}^6$ using the $\vee$ operator. Therefore, we can define the $6\times 6$ covariance matrix conveniently as \mbox{$\boldsymbol \Sigma_{\boldsymbol \epsilon} = \Covariance{\boldsymbol \epsilon}$}. We use the following adopted simplified notations from~\cite{forster2016manifold}:
\begin{equation} 
\small{
\label{eq:vectorized_maps}
\nonumber
\begin{array}{llll}
\Exp: \realnumbers^6 &\ni \twist &\rightarrow \exp(\twist[][\wedge])  &\in \SE(3) \\
\Log: \SE(3)         &\ni \homogeneous        &\rightarrow \log(\homogeneous)^\vee              &\in \realnumbers^6 .
\end{array}}
\end{equation}

The adjoint representation of a Lie group is a linear map that captures the non-commutative structure of the group. For $\SE(3)$, the matrix representation of the adjoint map~\cite{chirikjian2011stochastic} is
\begin{equation} \label{eq:adjoint}
\Adjoint[\homogeneous] = 
\begin{bmatrix}
\R & \zeros[3 \times 3] \\
\p[][\wedge] \R & \R \\
\end{bmatrix}.
\end{equation}
For any $\homogeneous \in \SE(3)$ and $\twist \in \realnumbers^6$, the adjoint map asserts
\begin{equation} 
\label{eq:adjoint_map}
\small
\nonumber (\Adjoint[\homogeneous] \twist)^\wedge = \homogeneous \twist[][\wedge] \homogeneous^{-1} \Rightarrow \Exp(\Adjoint[\homogeneous] \twist) = \homogeneous \Exp(\twist) \homogeneous^{-1}.
\end{equation}

\subsection{Retraction Map and Optimization on Manifold}
Given a retraction mapping and the associated manifold, we can optimize over the manifold by iteratively \emph{lifting} the cost function of our optimization problem to the tangent space, solving the re-parameterized problem, and then mapping the updated solution back to the manifold using the retraction~\cite{absil2009optimization}. For $\SE(3)$ we use its exponential map as the natural retraction, $\mathcal{R}_{\homogeneous}(\delta\twist) = \homogeneous \Exp ( \delta \twist)$, where $\delta \twist \in \realnumbers^6$ is the twist defined earlier in~\eqref{eq:twist}. Therefore, the retractions on the base and the contact poses are defined as
\begin{equation} \label{eq:retraction}%
\small
\begin{split} %
\X[i] &\leftarrow \X[i]\Exp(\delta \x[i]) \\%
\end{split}%
\quad \text{and} \quad%
\begin{split}%
\C[i] &\leftarrow \C[i]\Exp(\delta \c[i]). \\%
\end{split} %
\end{equation}%
These retractions are used to both update the state during optimization and to compute the Jacobians of the residuals.

\section{Problem Statement and Formulation}
\label{sec:problem}

In this section, we formulate the state estimation problem using the factor graph framework where independent sensor measurements can be incorporated by introducing additional factors based on the associated measurement models. The biped robot, shown in Figure \ref{fig:first_fig}, is equipped with a stereo camera, an IMU mounted on the torso, joint encoders, and a series of springs used for binary contact detection. Without loss of generality, we assume the IMU and camera are collocated with the base frame of the robot. 

%%%%%%%%%%%%%%%%%%%%%%%%%%%%%%%%%%%%%%%%%%%%%%%%
%\vspace{-2mm}
\subsection{State Representation}

The state variables (at time $t_i$) include the 3D body pose, $\homogeneous[WB](t) \in \SE(3)$, and velocity, $\linearVelocity[WB][W](t) \in \realnumbers^3$, in the world frame, the 3D contact pose in the world frame, $\homogeneous[WC](t) \in \SE(3)$, and the IMU bias, $\b(t) \triangleq \mathrm{vec}\left(\b[][g](t), \b[][a](t)\right) \in \realnumbers^6$, where $\b[][g](t) \in \realnumbers^3$ and $\b[][a](t) \in \realnumbers^3$ are the gyroscope and accelerometer biases, respectively. All together, the state at any time-step $i$ is a tuple as follows:
\begin{equation} \label{eq:state}
\resizebox{0.9\columnwidth}{!}{$ 
\Tcal_i \triangleq \left(\homogeneous[WB](t), \linearVelocity[WB][W](t), \homogeneous[WC](t), \b(t)\right) \triangleq \left(\X[i], \v[i], \C[i], \b[i]\right) $}
\end{equation}

\noindent Further, it is convenient to denote the trajectory of the state variables up to time-step $k$ by $\Xcal_k \triangleq \bigcup_{i=1}^{k} \Tcal_i$. Foot slip is the major source of drift in kinematic leg odometry. The inclusion of the contact pose in the state tuple allows for isolating the noise at the contact point.

%%%%%%%%%%%%%%%%%%%%%%%%%%%%%%%%%%%%%%%%%%%%%%%%%
%\vspace{-2mm}
\subsection{Factor Graph Formulation}

Let $\Lcal_{ij} \in \SE(3)$ be a perceptual loop-closure measurement relating poses at time-steps $i$ and $j$ ($j > i$) computed from an independent sensor, e.g.\@ using a point cloud matching algorithm. The forward kinematic measurements at time-step $i$ are denoted by $\Fcal_{i}$. The IMU and contact sensors provide measurements at higher frequencies. Between any two time-steps $i$ and $j$, we denote the set of all IMU and contact measurements by $\Ical_{ij}$ and $\Ccal_{ij}$, respectively. Let $\Kcal_k$ be the index set of time-steps (or keyframes) up to time-step $k$. We denote the set of all measurements up to time-step $k$ by $\Zcal_k \triangleq \{\Lcal_{ij}, \Ical_{ij},\Fcal_{i},\Ccal_{ij}\}_{i,j\in\Kcal_k}$.

By assuming the measurements are conditionally independent and are corrupted by additive zero mean white Gaussian noise, the posterior probability of the \emph{full SLAM} problem can be written as \mbox{$p(\Xcal_{k}|\Zcal_k) \propto p(\Xcal_0) p(\Zcal_k|\Xcal_k)$}, where
\begin{equation}
\resizebox{\columnwidth}{!}{$
  \nonumber p(\Zcal_k|\Xcal_k) = \prod_{i,j\in \Kcal_k} p(\Lcal_{ij}|\Xcal_j) p(\Ical_{ij}|\Xcal_j) p(\Fcal_{i}|\Xcal_i) p(\Ccal_{ij}|\Xcal_j).$}
\end{equation}

The \emph{Maximum-A-Posteriori} (MAP) estimate of $\Xcal_{k}$ can be computed by solving the following optimization problem:
\begin{equation}
  \label{eq:fullslam}
  \nonumber \underset{\Xcal_k}{\minimize} \ -\log p(\Xcal_k|\Zcal_k)
\end{equation}
in which due to the noise assumption mentioned earlier is equivalent to the following nonlinear least-squares problem:
\begin{equation}
\label{eq:map_nls}
  \small
  \begin{split}
  \nonumber \underset{\Xcal_k}{\minimize} \ &\lVert \r[0] \rVert^2_{\Cov[0]} + \sum_{i,j\in\Kcal_k} \lVert \r[\Lcal_{ij}] \rVert^2_{\Cov[\Lcal_{ij}]} + \sum_{i,j\in\Kcal_k} \lVert \r[\Ical_{ij}] \rVert^2_{\Cov[\Ical_{ij}]}\\ 
  & + \sum_{i\in\Kcal_k} \lVert \r[\Fcal_{i}] \rVert^2_{\Cov[\Fcal_{i}]} + \sum_{i,j\in\Kcal_k} \lVert \r[\Ccal_{ij}] \rVert^2_{\Cov[\Ccal_{ij}]}
  \end{split}
\end{equation}
where $\r[0]$ and $\Cov[0]$ represent the prior over the initial state and serves to anchor the graph, $\r[\Lcal_{ij}]$, $\r[\Ical_{ij}]$, $\r[\Fcal_{i}]$, $\r[\Ccal_{ij}]$ are the residual terms associated with the loop closure, IMU, forward kinematic, and contact measurements respectively, i.e.\@ the error between the measured and predicted values given the state, and $\Cov[\Lcal_{ij}]$, $\Cov[\Ical_{ij}]$, $\Cov[\Fcal_{i}]$, $\Cov[\Ccal_{ij}]$ are the corresponding covariance matrices.

%%%%%%%%%%%%%%%%%%%%%%%%%%%%%%%%%%%%%%%%%%%%%%%%%

\section{Forward Kinematics}
\label{sec:fk}

Forward kinematics refers to the process of computing the relative pose transformation between two frames of a multi-link system. Each individual joint displacement describes how the child link moves with respect to the parent one. This joint displacement can either be an angle (revolute joints) or a distance (prismatic joints). 

\begin{figure}[t]
  \vspace{2mm}
  \centering 
  \includegraphics[width=0.91\columnwidth]{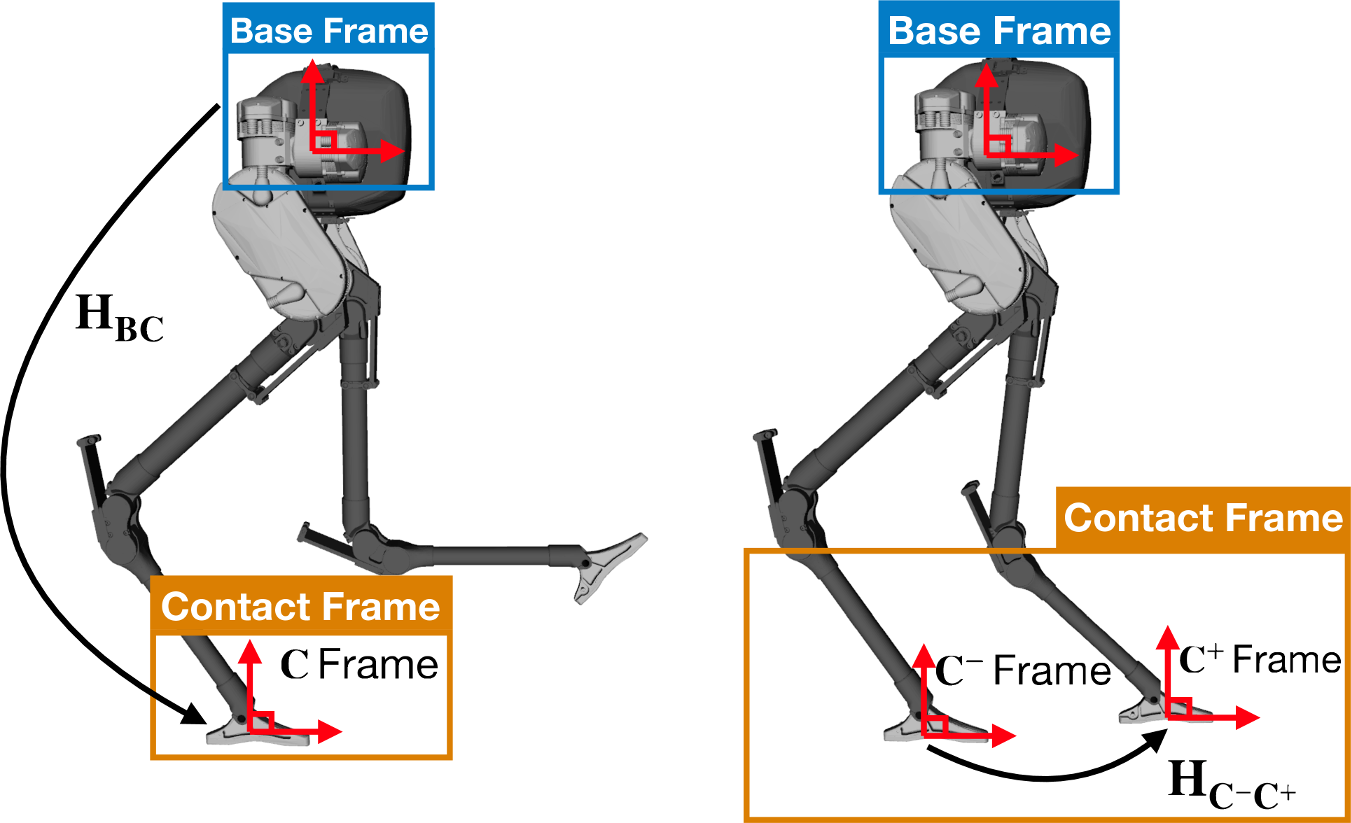}
  \caption{In this paper, we refer to two separate forward kinematics functions. The pose of the current contact frame relative to the base frame is denoted by $\homogeneous[BC]$. When the robot has multiple points of contact with the environment, it is possible to transfer this contact from from one frame to another. This transfer of contact is captured by the homogeneous transform $\homogeneous[C\textsuperscript{-}C\textsuperscript{+}]$.}
  \squeezeup\squeezeup
  \label{fig:fk}
\end{figure}

Let $\encoders \in \realnumbers^N$ denote the vector of joint displacements for a general robot. Without loss of generality, we define a \emph{base frame} on the robot, denoted B, that is assumed to be collocated with both the IMU and the camera frames. When the robot is in contact with the static environment, we can also define a \emph{contact frame}, denoted C, on the robot at the point of contact. The homogeneous transformation between the base frame and the contact frame is defined by:
\begin{equation} \label{eq:forward_kinematics_base_contact}
\homogeneous[BC](\encoders) \triangleq 
\begin{bmatrix}
\orientation[BC](\encoders)  & \position[BC][B](\encoders)  \\
\zeros[1 \times 3] & 1
\end{bmatrix},
\end{equation}
where $\orientation[BC](\encoders)$ and $ \position[BC][B](\encoders)$ denote the relative orientation and position of the contact frame with respect to the base frame.

When there are two points in contact with the static environment, it is possible to ``transfer'' the contact frame from one point to the other (shown in Figure \ref{fig:fk}). Let $\text{C}^-$ denote the old contact frame and $\text{C}^+$ denote the new contact frame. Then, the homogeneous transformation between the old frame and the new frame is defined by:
\begin{equation} \label{eq:forward_kinematics_contact1_contact2}
\homogeneous[C\textsuperscript{-}C\textsuperscript{+}](\encoders) \triangleq
\begin{bmatrix}
\orientation[C\textsuperscript{-}C\textsuperscript{+}](\encoders) & \position[C\textsuperscript{-}C\textsuperscript{+}][C\textsuperscript{-}](\encoders) \\
\zeros[1 \times 3] & 1
\end{bmatrix},
\end{equation}
where $\orientation[C\textsuperscript{-}C\textsuperscript{+}](\encoders)$ and $\position[C\textsuperscript{-}C\textsuperscript{+}][C\textsuperscript{-}](\encoders)$ denote the relative orientation and position of the new contact frame with respect to the old contact frame. 
  
%%%%%%%%%%%%%%%%%%%%%%%%%%%%%%%%%%%%%%%%%%%%%%%%%%%%%
\subsection{Measurements}

We assume the robot's joints are equipped with a set of joint encoders that can measure the joint displacement. These encoder measurements, $\encodersM$, are assumed to be corrupted with Gaussian white noise. This is an explicit measurement coming from physical sensors located on the robot.
\begin{equation} \label{eq:encoder_measurements}
\encodersM(t) = \encoders(t) + \noise[][\alpha](t), \quad \noise[][\alpha](t) \sim \mathcal{N}(\zeros[N \times 1], \Cov[][\alpha](t)) \\
\end{equation}

The geometric (or manipulator) Jacobian, denoted $\Jacobian(\encoders)$, provides a method for computing the angular and linear velocity of an end-effector given the vector of joint velocities~\cite{murray1994mathematical}. In a similar manner, we can use the Jacobian to map incremental angle displacements to changes in the end-effector pose. Let $\Jacobian[BC](\encoders)$ denote the body manipulator Jacobian of the forward kinematics function \eqref{eq:forward_kinematics_base_contact}. Then, the following relationship holds: 
\begin{equation} \label{eq:jacobian_fk_noise}
{}_\text{C}\twist[\text{BC}] \, \delta t = 
\Jacobian[BC](\encoders) \, \delta \encoders
\end{equation}
where $\delta \encoders$ and $\delta t$ are incremental encoder and time quantities and ${}_\text{C}\twist[\text{BC}]$ denotes the vector of angular and linear velocities of the contact frame due to $\delta \encoders$ (measured in the contact frame). We can use this relationship \eqref{eq:jacobian_fk_noise} to factor out the noise from the forward kinematics equations. Up to a first order approximation, \eqref{eq:forward_kinematics_base_contact} can be factored as: 
\begin{equation} \label{eq:forward_kinematics_factor_noise}
\resizebox{0.85\columnwidth}{!}{$ 
\homogeneous[BC](\encoders(t)) \approx \homogeneous[BC](\encodersM(t)) \;  \Exp\left(-\Jacobian[BC](\encodersM(t)) \noise[][\alpha](t) \right).
$}
\end{equation}
A similar approximation can be found for \eqref{eq:forward_kinematics_contact1_contact2}, albeit with a different Jacobian. 
\begin{equation} \label{eq:forward_kinematics_switch_factor_noise}
\resizebox{0.85\columnwidth}{!}{$
\homogeneous[C\textsuperscript{-}C\textsuperscript{+}](\encoders(t)) \approx \homogeneous[C\textsuperscript{-}C\textsuperscript{+}](\encodersM(t)) \; \Exp\left(- \Jacobian[C\textsuperscript{-}C\textsuperscript{+}](\encodersM(t)) \noise[][\alpha](t) \right)
$}
\end{equation}

\begin{remark}
In general, the manipulator Jacobian can be derived as spatial or body Jacobian~\cite{murray1994mathematical}, and based on this choice, the noise can appear on the left or right side of the rotation/rigid body transformation, respectively.
\end{remark}

%%%%%%%%%%%%%%%%%%%%%%%%%%%%%%%%%%%%%%%%%%%%%%%%%%%%%
\subsection{Forward Kinematic Factor}
The goal of this section is to derive a general \emph{forward kinematic factor} that can be used in the factor graph framework. This will be a unary factor that relates two poses through the forward kinematics equations while accounting for encoder noise. We derive it here for relating the base and contact frames.

The orientation and position of the contact frame with respect to the world frame are given by:
\begin{equation} 
\homogeneous[WC](t) = \homogeneous[WB](t) \; \homogeneous[BC](\encoders(t)).
\end{equation}
Substituting in the state variables at time $t_i$ \eqref{eq:state}, the forward kinematic equation \eqref{eq:forward_kinematics_base_contact} yields:
\begin{equation} 
\C[i] = \X[i] \; \homogeneous[BC](\encodersM[i] - \noise[i][\alpha]).
\end{equation}
We can now use the first order approximation \eqref{eq:forward_kinematics_factor_noise} to factor out the encoder noise to give the following expression:
\begin{equation} 
\C[i] = \X[i] \; \homogeneous[BC](\encodersM[i]) \; \Exp\left( - \Jacobian[BC](\encodersM[i]) \noise[i][\alpha] \right).
\end{equation}

Defining the zero-mean white Gaussian forward kinematic noise term, $\delta \c[i] \triangleq \Jacobian[BC](\encodersM[i]) \noise[i][\alpha]$, allows us to write out the \emph{forward kinematic measurement model}:
\begin{equation} \label{eq:FK_measurement_model}
\homogeneous[BC](\encodersM[i]) = \X[i][-1] \; \C[i] \; \Exp(\delta \c[i] ),
\end{equation}
where the forward kinematic noise is characterized by $ \delta \c[i] \sim \mathcal{N}(\zeros[6 \times 1],\Cov[\Fcal_{i}])$. The residual error is defined in the tangent space and can be written as: 
\begin{equation} 
\r[\Fcal{i}] = \Log\left( \C[i][-1]\; \X[i] \; \homogeneous[BC](\encodersM[i]) \right).
\end{equation}
The covariance is computed through the following linear transformation:
\begin{equation}
\Cov[\Fcal_{i}] = 
\Jacobian[BC](\encodersM[i]) \; \Cov[i][\alpha] \; \Jacobian[BC]^\transpose(\encodersM[i]),
\end{equation}
where $\Jacobian[BC](\encodersM[i])$ is the body manipulator Jacobian evaluated at the current encoder measurements and $\Cov[i][\alpha]$ denotes the encoder covariance matrix at time $t_i$. 

\section{Hybrid Contact Preintegration}
\label{sec:contact_preintegration}

% \subsection{Definitions}

A continuous hybrid dynamical system, $\Hcal$, can be defined with a continuous dynamics function, $f(\cdot)$, a discrete transition map, $\Delta(\cdot)$, and a switching surface, $\Scal$ \cite{westervelt2007feedback}. Trajectories of the hybrid dynamical system evolve according to the continuous dynamics, until the switching surface it hit. At those moments, the state gets mapped through the discrete transition map, after which the trajectory continues according to the continuous dynamics again. The general form of this system can be expressed as follows:
\begin{equation} \label{eq:continuous_hybrid_system}
\mathcal{H} : 
\begin{cases}
	\begin{alignedat}{2}
	\dot{x}(t) &= f(x,t) \qquad (x^-,t^-) \not\in \Scal \\
	x^+ &= \Delta(x^-) \qquad (x^-,t^-) \in \Scal. \\
    \end{alignedat}
\end{cases} 
\end{equation}
As long as the number of contact points is greater than or equal to one, the switching contact frame dynamics can be modeled as a hybrid system:
\begin{equation} \label{eq:continuous_hybrid_contact_model}
    \mathcal{H} : 
    \begin{cases}
        \begin{aligned}
            \homogeneousDot[WC](t) &= \homogeneous[WC](t) \; \left({}_\text{C}\twist[\text{WC}](t)\right)^\wedge \quad &&t^- \not\in \Scal \\
            \homogeneous[WC\textsuperscript{+}] &= \homogeneous[WC\textsuperscript{-}] \; \homogeneous[C\textsuperscript{-}C\textsuperscript{+}] \quad &&t^- \in \Scal, \\
        \end{aligned}
    \end{cases} 
\end{equation}
where the switching surface, $\Scal$ is simply modeled as the set of all times where contact is switched from one frame to another. Since the sensor measurements are coming in at discrete time-steps, we perform Euler integration from time $t$ to $t+\Delta t$ to discretize the continuous hybrid contact dynamics \eqref{eq:continuous_hybrid_contact_model}, forming the following discrete hybrid system:
\begin{equation} \label{eq:discrete_hybrid_contact_model_general}
\mathcal{H} : 
\begin{cases}
	\begin{aligned}
        \homogeneous[WC](t + \Delta t) &= \homogeneous[WC](t) \; \Exp\left( {}_\text{C}\twist[\text{WC}](t) \Delta t \right) &&t^- \not\in \Scal \\
        \homogeneous[WC\textsuperscript{+}] &= \homogeneous[WC\textsuperscript{-}] \; \homogeneous[C\textsuperscript{-}C\textsuperscript{+}] &&t^- \in \Scal.\\
    \end{aligned}
\end{cases} 
\end{equation}

Physically, the continuous dynamics function, $f$, describes how a single contact frame moves over time while contact is maintained. When a new contact frame is detected, the new contact pose can be computed by applying the transition map, $\Delta(\cdot)$, which describes the homogeneous transformation between the old and new contact frames.

\vspace{-2mm}
\subsection{Measurements}
The angular and linear velocity of the contact point is an implicit measurement that is \textit{inferred} through a binary contact sensor: specifically, when this sensor indicates contact, the pose of the contact frame is \textit{assumed to remain fixed with respect to the world frame}, i.e. the measured velocities are zero. In order to accommodate potential contact slip, this measured twist is modeled to be corrupted with white Gaussian noise, namely
\begin{equation} 
\label{eq:contact_measurements}
\resizebox{\columnwidth}{!}{$
\nonumber {}_\text{C}\twistM[\text{WC}](t) = \zeros[6 \times 1] = {}_\text{C}\twist[\text{WC}](t)  + \noise[][\xi](t), \hspace{2mm} \noise[][\xi](t) \sim \mathcal{N}(\zeros[6 \times 1], \Cov[][\xi](t)).$}
\end{equation}
Using the state variables \eqref{eq:state}, forward kinematics definition \eqref{eq:forward_kinematics_contact1_contact2}, encoder measurements \eqref{eq:encoder_measurements}, and contact measurements, the hybrid contact dynamics \eqref{eq:discrete_hybrid_contact_model_general} can be written as:
\begin{equation} \label{eq:discrete_hybrid_contact_model}
    \small
    \mathcal{H} : 
    \begin{cases}
        \begin{aligned}
        \C[k+1] &= \C[k] \; \Exp\left( -\noise[k][\xi d] \Delta t \right) \quad &&t_k^- \not\in \Scal \\
        \C[][+] &= \C[][-] \; \homogeneous[C\textsuperscript{-}C\textsuperscript{+}](\encodersM[k] - \noise[k][\alpha]) \quad &&t_k^- \in \Scal \\
        \end{aligned}
    \end{cases} 
\end{equation}
where $\noise[k][vd]$, the discrete time contact noise, is computed using the sample time, $\small \Covariance{\noise[][\xi d](t)} = \dfrac{1}{\Delta t} \Covariance{\noise[][\xi](t)}$.

\begin{remark}
% The equations defining the forward kinematics function, 
$\homogeneous[C\textsuperscript{-}C\textsuperscript{+}](\encoders[k])$ will depend on the specific contact frames, $C^-$ and $C^+$, at time $t_k$. For example, the forward kinematics function to switch from left foot contact to a right foot contact will be different than the one used to switch from right foot to left foot. 
%Therefore, when implementing the derived factor, the user needs to determine the specific frames involved in the contact transfer in order to select the appropriate kinematics function. 
\end{remark}

%%%%%%%%%%%%%%%%%%%%%%%%%%%%%%%%%%%%%%%%%%%%%%%%%%
\subsection{Preintegrating Contact Pose}
The goal of this section is to formulate a general \emph{hybrid preintegrated contact factor} which relates the contact pose at $t_i$ to the contact pose at $t_j$. The hybrid nature of this factor comes from the potential switching of contact that occurs naturally in legged locomotion. Following the work on IMU preintegration theory \cite{lupton2012visual,forster2016manifold}, we preintegrate the high-frequency contact measurements to prevent unnecessary computation allowing efficient implementation of the factor.

\begin{figure}[t]
  \vspace{1.5mm}
  \centering 
  \includegraphics[width=0.95\columnwidth]{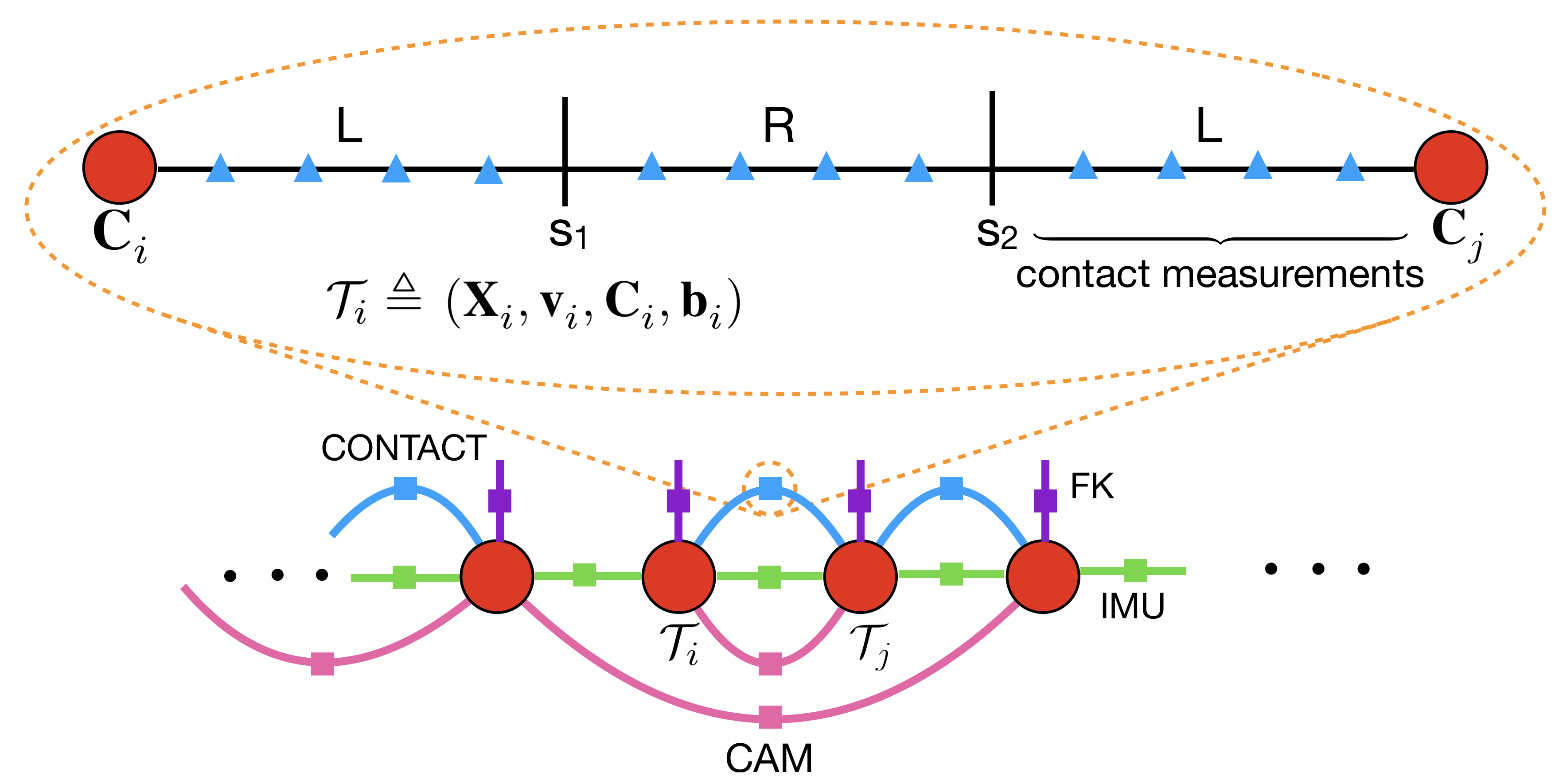}
  \caption{In the factor graph framework, the robot's state along a discretized trajectory denoted by red circles. Each independent sensor measurement is a factor denoted by lines that constraints the state at separate time-steps. The proposed hybrid contact factor (shown on top) allows preintegration of high-frequency contact data through an arbitrary number of contact switches. In this example, there are two contact switches, where the robot moves from left-stance (L) to right-stance (R), then back to left stance.}
  \squeezeup%\squeezeup
  \label{fig:example_switches}
\end{figure}

Let $\Scal$ denote the sequence of all time indices associated with contact switches, so that each $s_i \in \Scal$ represents a single time index where a contact switch occurs. To integrate the hybrid contact dynamics model \eqref{eq:discrete_hybrid_contact_model} through these contact switches, we integrate the up to the next contact switch time $t_{s_i}$, apply the transition map $\Delta(\cdot)$, and continue the integration. For the following derivations, a concrete example (depicted in Figure \ref{fig:example_switches}) is used to make the derivation easier to follow. We later extend the derivation to an arbitrary number of contact switches. For the example shown in Figure \ref{fig:example_switches}, there are two contact switches between times $t_i$ and $t_j$, i.e., $\lvert \Scal \rvert = 2$. Integrating the discrete hybrid contact model \eqref{eq:discrete_hybrid_contact_model} from $t_i$ to $t_j$ yields:
\begin{equation*}
\resizebox{\hsize}{!}{$
\begin{split}
&\C[j] = \C[i] \left( \prod_{k=i}^{s_1-1} \Exp( -\noise[k][\xi d] \Delta t) \right) \homogeneous[C\textsuperscript{-}C\textsuperscript{+}] (\encodersM[s_1] - \noise[s_1][\alpha])\\
& \left( \prod_{k=s_1}^{s_2-1} \Exp( -\noise[k][\xi d] \Delta t) \right) \homogeneous[C\textsuperscript{-}C\textsuperscript{+}](\encodersM[s_2] - \noise[s_2][\alpha]) \left( \prod_{k=s_2}^{j-1} \Exp( -\noise[k][\xi d] \Delta t) \right). \\
\end{split}$}
\end{equation*}
All noise terms can be shifted to the right by using \eqref{eq:forward_kinematics_switch_factor_noise} to factor the noise  from the forward kinematics term and the adjoint relation \eqref{eq:adjoint} to shift the measured kinematics to the left.
\begin{equation}
\small
\begin{split}
&\C[j] = \C[i] \; \homogeneous[C\textsuperscript{-}C\textsuperscript{+}] (\encodersM[s_1]) \; \homogeneous[C\textsuperscript{-}C\textsuperscript{+}] (\encodersM[s_2]) \\ 
& \left( \prod_{k=i}^{s_1-1} \Exp( -\Adjoint[\homogeneous[C\textsuperscript{-}C\textsuperscript{+}]^{-1}(\encodersM[s_2])] \Adjoint[\homogeneous[C\textsuperscript{-}C\textsuperscript{+}]^{-1}(\encodersM[s_1])] \noise[k][\xi d] \Delta t) \right)  \\
&\Exp(-\Adjoint[\homogeneous[C\textsuperscript{-}C\textsuperscript{+}]^{-1}(\encodersM[s_2])] \Jacobian[C\textsuperscript{-}C\textsuperscript{+}](\encodersM[s_1]) \noise[s_1][\alpha]) \\ 
& \left( \prod_{k=s_1}^{s_2-1} \Exp( -\Adjoint[\homogeneous[C\textsuperscript{-}C\textsuperscript{+}]^{-1}(\encodersM[s_2])] \noise[k][\xi d] \Delta t) \right)\\
&\Exp( - \Jacobian[C\textsuperscript{-}C\textsuperscript{+}](\encodersM[s_2]) \noise[s_2][\alpha]) \left( \prod_{k=s_2}^{j-1} \Exp( -\noise[k][\xi d] \Delta t) \right) \\
\end{split}
\end{equation}
After multiplying both sides by $\C[i][-1]$, we arrive at a relative pose expression that is independent of states $\Tcal_i$ and $\Tcal_j$.
\begin{equation} \label{eq:relative_orientation}
 \Delta \C[ij] \triangleq \C[i][-1] \C[j] = \Delta \CM[ij] \Exp(-\delta \c[ij]) \\
\end{equation} 
where $\Delta \CM[ij] \triangleq \prod_{n=1}^{|\mathcal{S}|}  \homogeneous[C\textsuperscript{-}C\textsuperscript{+}] (\encodersM[s_n])$ represents the \emph{hybrid preintegrated contact measurement}, and $\Exp(-\delta \c[ij])$ groups all of the noise terms together. This noise term is a product of multiple small rigid body transformations. Therefore, it can be approximated as a summation in the tangent space through iterative use of the Baker-Campbell-Hausdorff
(BCH) formula~\cite{chirikjian2011stochastic} (while keeping only the first order terms). 
\begin{equation} \label{eq:approximate_preintegrated_contact_noise}
\begin{split}
\small
\nonumber \delta &\c[ij] \approx \sum_{k=i}^{s_1-1} \Adjoint[\homogeneous[C\textsuperscript{-}C\textsuperscript{+}]^{-1}(\encodersM[s_2])] \Adjoint[\homogeneous[C\textsuperscript{-}C\textsuperscript{+}]^{-1}(\encodersM[s_1])] \noise[k][\xi d] \Delta t \\
& + \Adjoint[\homogeneous[C\textsuperscript{-}C\textsuperscript{+}]^{-1}(\encodersM[s_2])] \Jacobian[C\textsuperscript{-}C\textsuperscript{+}](\encodersM[s_1]) \noise[s_1][\alpha] + \sum_{k=s_1}^{s_2-1} \Adjoint[\homogeneous[C\textsuperscript{-}C\textsuperscript{+}]^{-1}(\encodersM[s_2])] \noise[k][\xi d] \Delta t \\
& + \Jacobian[C\textsuperscript{-}C\textsuperscript{+}](\encodersM[s_2]) \noise[s_2][\alpha] + \sum_{k=s_2}^{j-1} \noise[k][\xi d] \Delta t
\end{split} 
\end{equation}
The \emph{hybrid preintegrated contact noise}, $\delta \c[ij]$, is a summation of zero-mean Gaussian terms, and is therefore also zero-mean and Gaussian. It is possible to generalize this noise expression to an arbitrary number of contact switches, however, it becomes much simpler to do so when looking at the iterative propagation form in the following section.

\subsection{Iterative Propagation}
It is possible to write both the preintegrated contact measurements, $\Delta \CM[ij]$, and the preintegrated contact noise, $\delta \c[ij]$, in iterative update forms. This allows the terms to be updated as contact and encoder measurements are coming in. In addition, this form simplifies the expressions and allows for the covariance to be conveniently computed. The following proposition generalizes the hybrid preintegrated contact pose and noise iterative propagation to an arbitrary number of contact switches. The proof is given in the supplementary material.

\begin{proposition}[Iterative Propagation of Hybrid Contact Process~\cite{iros_supplementary}] \label{prop:iterative_propagation}
Between any two time-steps $\normalfont t_i$ and $\normalfont t_j$ such that $j > i$, starting with $\normalfont\Delta \CM[ii] = \I[4]$ and $\normalfont\delta \c[ii] = \zeros[3 \times 1]$, the hybrid preintegrated contact measurement, $\normalfont\Delta \CM[ij]$, and noise, $\normalfont\delta \c[ij]$, for an arbitrary number of contact switches can be computed iteratively using the following hybrid systems:
\normalfont
\begin{equation}
    \small
    \tilde{\Hcal}:
    \begin{cases}
        \begin{aligned}
            \Delta \CM[ik+1] &= \Delta \CM[ik] \quad &&t_k^- \not\in \Scal \\
            \Delta \CM[ik][+] &= \Delta \CM[ik][-] \; \homogeneous[C\textsuperscript{-}C\textsuperscript{+}](\encodersM[k])  \quad &&t_k^- \in \Scal \\
        \end{aligned}
    \end{cases}
\end{equation}
\begin{equation} \label{eq:hybrid_contact_noise_model}
    \small
    \delta \Hcal:
    \begin{cases}
        \begin{aligned}
            \delta \c[ik+1] &= \delta \c[ik] + \noise[k][\xi d] \Delta t \quad &&t_k^- \not\in \Scal \\
            \delta \c[ik][+] &= \Adjoint[\homogeneous[C\textsuperscript{-}C\textsuperscript{+}]^{-1}(\encodersM[k])] \delta \c[ik][-] + \Jacobian[C\textsuperscript{-}C\textsuperscript{+}](\encodersM[k]) \noise[k][\alpha] \quad &&t_k^- \in \Scal
        \end{aligned}
    \end{cases}
\end{equation}
\end{proposition}
\squeezeup \noindent
An abstract representation of these hybrid systems is shown in Figure \ref{fig:HybridModelLieGroup}.

\begin{figure}[t]
  \vspace{1.5mm}
  \centering
  \includegraphics[width=1.00\columnwidth]{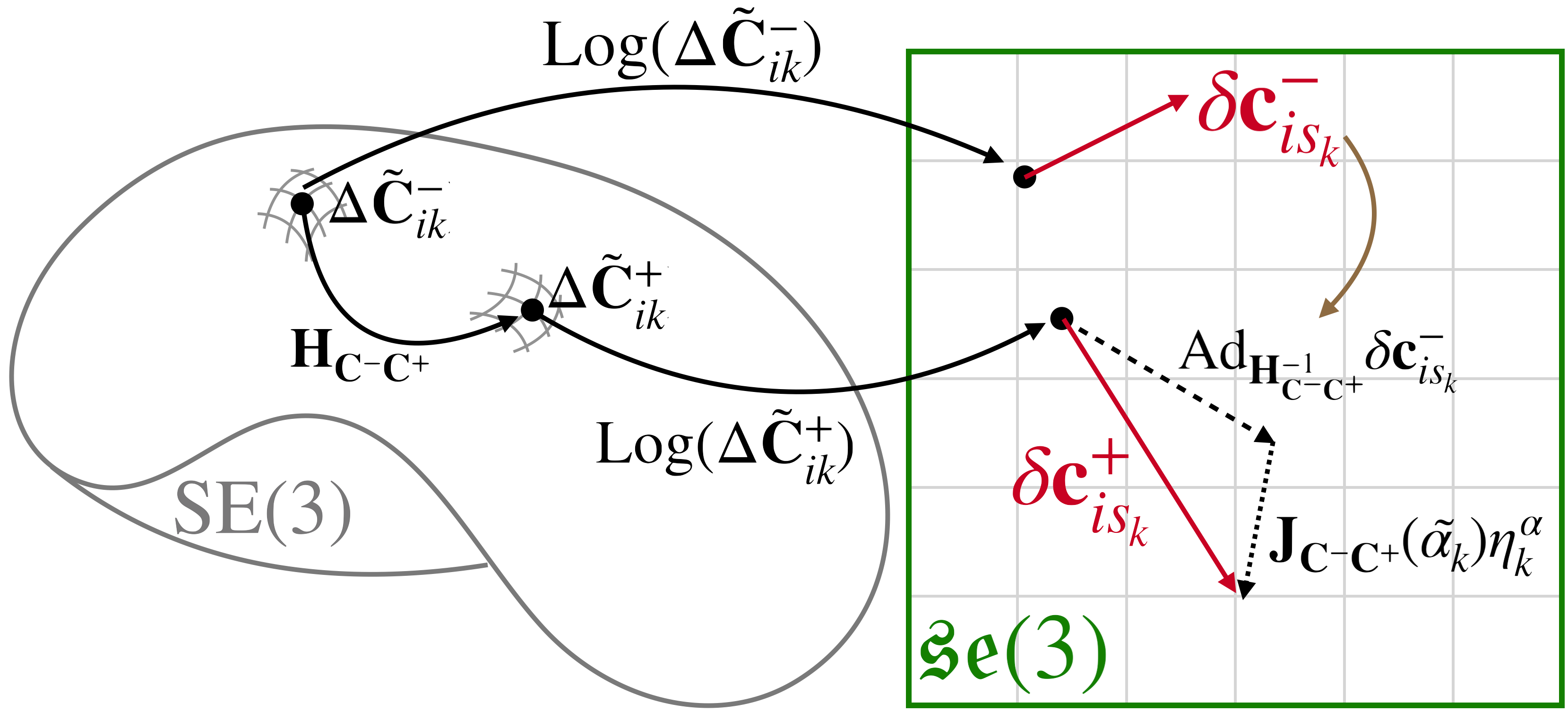}
  \caption{When a contact switch occurs, the relative contact pose, $\scriptsize \Delta \CM[ik][-]$, gets mapped from one point in $\SE(3)$ to another point, $\scriptsize \Delta \CM[ik][+]$, on the manifold. The contact noise, $\scriptsize \delta \c[ik][-]$, is represented in the tangent space, $\mathfrak{se}(3)$, and is mapped from the tangent space of $\scriptsize \Delta \CM[ik][-]$ to the tangent space of $\scriptsize \Delta \CM[ik][+]$ through the use of the adjoint map of the forward kinematics transformation, $\scriptsize \homogeneous[C\textsuperscript{-}C\textsuperscript{+}]$. However, due to noisy encoders, an addition noise term (computed using the manipulator Jacobian) has to be added to compute $\scriptsize \delta \c[ik][+]$. }
  \label{fig:HybridModelLieGroup}
  \squeezeup\squeezeup
\end{figure}
   
%%%%%%%%%%%%%%%%%%%%%%%%%%%%%%%%%%%%%%%%%%%
%\vspace{-5mm}
\subsection{Rigid Contact Factor}
The relative contact pose expression \eqref{eq:relative_orientation} can be used to define the \emph{hybrid preintegrated contact measurement model}:
\begin{equation} \label{eq:preintegrated_contact_measurement_model}
\Delta \CM[ij] = \C[i][-1]\C[j] \Exp(\delta\c[ij]),
\end{equation} 
where the preintegrated contact noise is characterized by $ \delta\c[ij] \sim \mathcal{N}(\zeros[6 \times 1], \Cov[\Ccal_{ij}])$. In the factor graph framework, this hybrid preintegrated contact model represents a binary factor that relates the contact frame pose over consecutive time steps. The residual error is defined in the tangent space, and can be written as: 
\begin{equation} 
\r[\Ccal_{ij}] = \Log\left( \C[j][-1] \C[i] \Delta \CM[ij](\encodersM[i])\right).
\end{equation}
The covariance is computed using the hybrid contact noise model \eqref{eq:hybrid_contact_noise_model}, starting with $\Cov[\Ccal_{ii}] = \zeros[6 \times 6]$:
\begin{equation}
  \small
  \begin{cases}
    \begin{aligned}
    \Cov[\Ccal_{ik+1}] &= \Cov[\Ccal_{ik}] + \Cov[k][\xi] \; \Delta t \quad &&t_k^- \not\in \Scal \\  
    \Cov[\Ccal_{ik}][+] &= \Adjoint[\homogeneous[C\textsuperscript{-}C\textsuperscript{+}]^{-1} \;(\encodersM[k])] \Cov[\Ccal_{ik}][-] \; \Adjoint[\homogeneous[C\textsuperscript{-}C\textsuperscript{+}]^{-1}(\encodersM[k])]^\transpose \quad &&t_k^- \in \Scal \\
    &\qquad\quad + \Jacobian[C\textsuperscript{-}C\textsuperscript{+}](\encodersM[k]) \; \Cov[k][\alpha] \; \Jacobian[C\textsuperscript{-}C\textsuperscript{+}]^\transpose(\encodersM[k]).
    \end{aligned}
  \end{cases}
\end{equation}

\section{Experimental Results}
\label{sec:results}

\begin{figure*}[t]
  \centering  
  \subfloat{\includegraphics[width=0.6666\columnwidth,trim={0.5cm 1.5cm 1cm 0.5cm},clip]{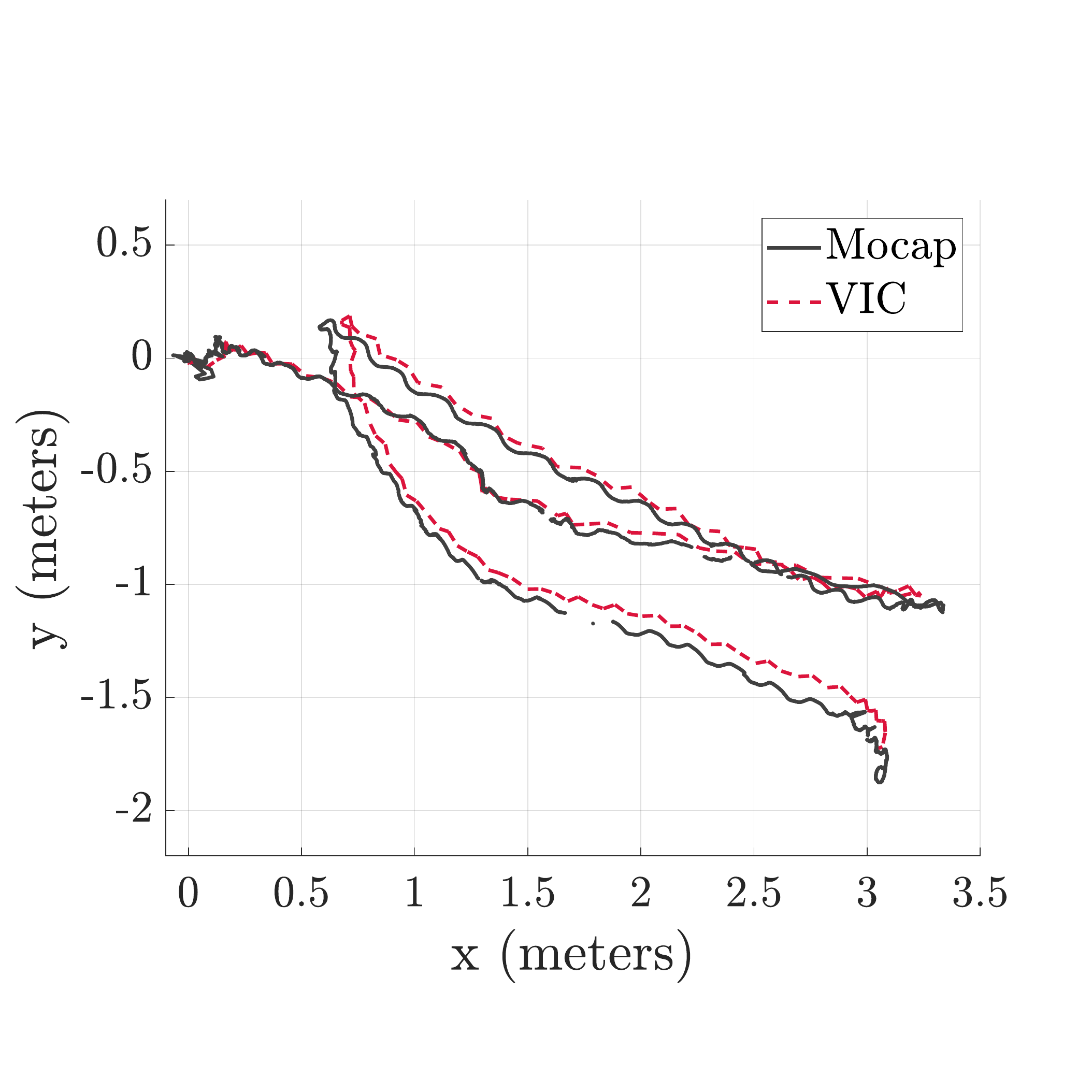}}
  \subfloat{\includegraphics[width=0.6666\columnwidth,trim={0.5cm 1.5cm 1cm 0.5cm},clip]{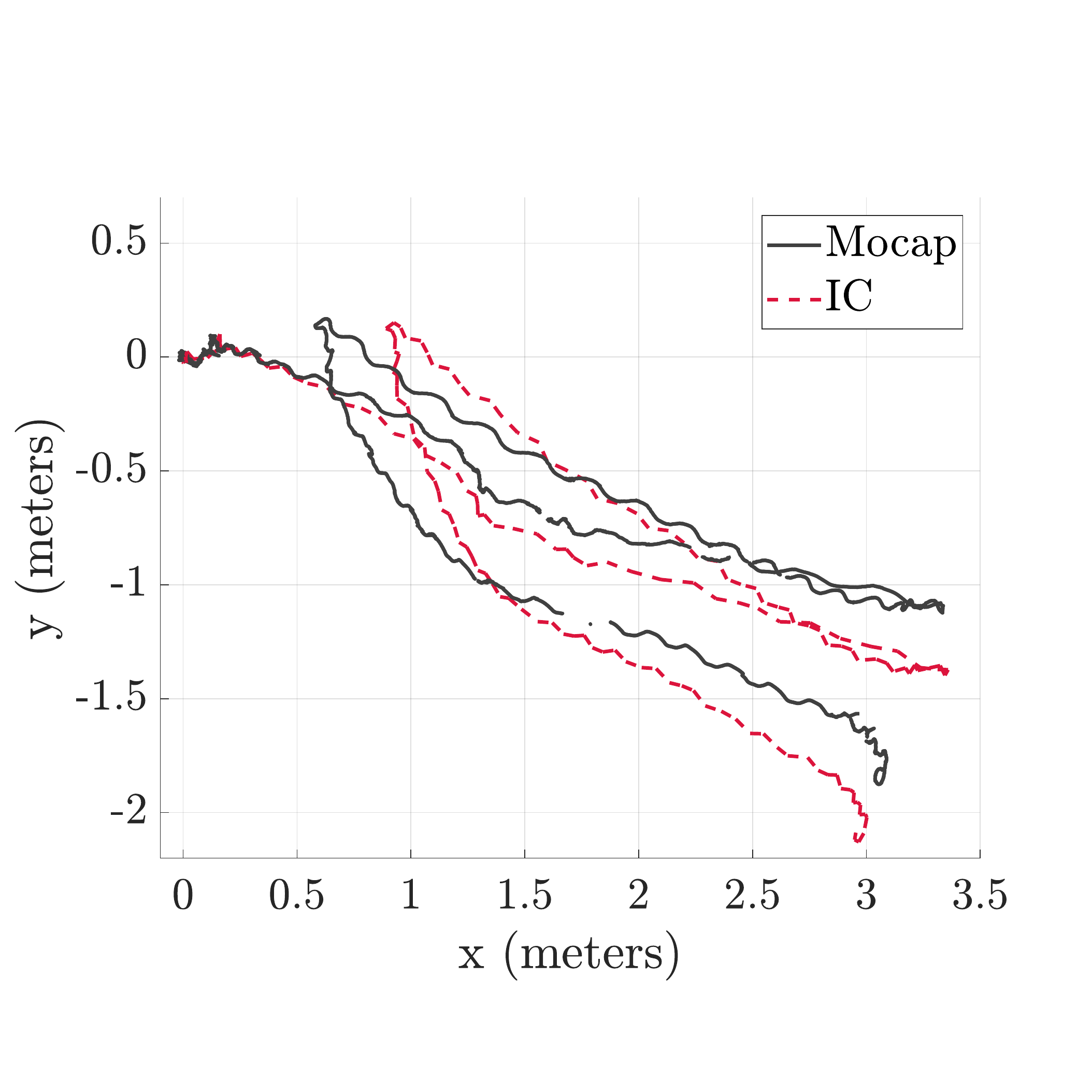}}
  \subfloat{\includegraphics[width=0.6666\columnwidth,trim={0.5cm 1.5cm 1cm 0.5cm},clip]{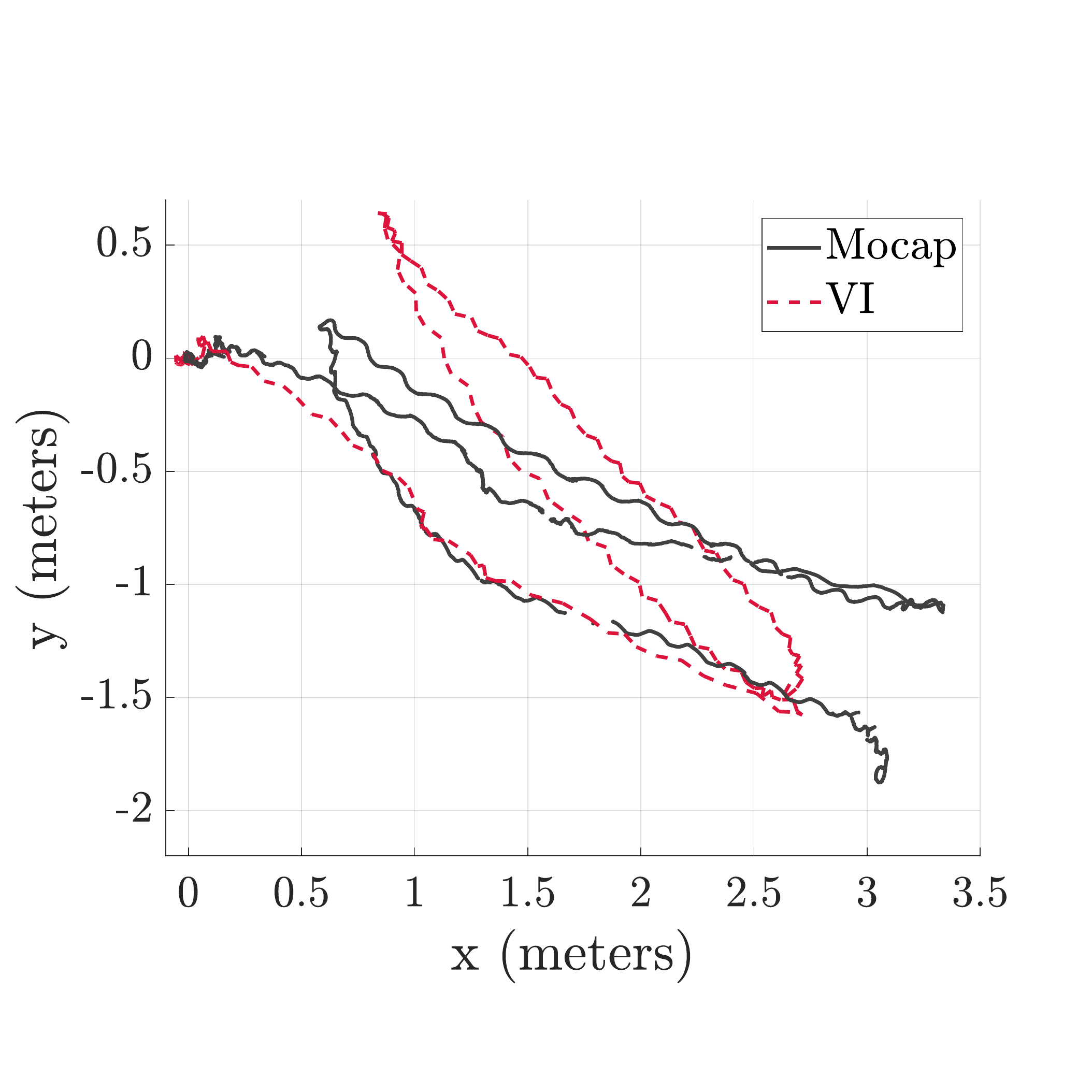}}
  \caption{The odometry results from a 60 second walking experiment using a Cassie-series robot. The Visual-Inertial-Contact (VIC) odometry outperformed both the Inertial-Contact (IC), and the Visual-Inertial (VI) odometry. ``Ground-truth'' data was collected from a Vicon motion capture system. It is important to note that no loop-closures are being performed, which helps to explain the relatively poor odometry from VI. The video of this experiment is provided at \href{https://youtu.be/WDPhdl5g2MQ}{https://youtu.be/WDPhdl5g2MQ}.}
  \squeezeup
  \label{fig:estimated_trajectory}
\end{figure*}

We now present experimental evaluations of the proposed factors. In the first experiment, we compare three odometry systems composed by Visual-Inertial-Contact (VIC), Inertial-Contact (IC), and Visual-Inertial (VI) factors. Whenever the contact factor is used, it is assumed that forward kinematic factor is also available; therefore, it is not explicitly mentioned for brevity. In the second experiment, we study the effect of losing visual data for a period of time to see how the contact factor can constrain the graph in the absence of a reliable vision system. In the third experiment, we will evaluate the use of \emph{terrain factors}, where loop-closures can be added to the graph through contact frame poses.

% factors: imu factor, vision factor, forward kinematic factor. All of the factors were implemented in GTSAM~\cite{dellaert2012factor} and were all built into GTSAM~4~\cite{forster2016manifold} along with iSAM2 as a solver~\cite{kaess2012isam2}. We compare two experiments as below: \bhg{two names of experiments}. \mgj{Citing the new IMU factor too?}

\subsection{Experimental Setup}
All experiments are done on a Cassie-series robot designed by Agility Robotics, which has 20 degrees of freedom, 10 actuators, joint encoders, and a VN-100 IMU. A Multisense S7 stereo camera, which contains a separate IMU, is mounted on the top of the Cassie robot as shown in Fig.~\ref{fig:first_fig}. The robot also has four springs (two on each leg) that can be used as binary contact sensors by thresholding the spring deflection measurements. The Cassie robot has two computers which run MATLAB Simulink Real-Time (for the controller) and Ubuntu (for the estimator) respectively. We use the Robot Operating System (ROS)~\cite{quigley2009ros} with the User Datagram Protocol (UDP) to communicate sensor data between the two computers. We also integrate the time synchronization algorithm in~\cite{olson2010} into our system to ensure all sensory data is synchronized.

The motion capture system developed by Vicon is used as a proxy for ground truth trajectories. The setup consists of 17 motion capture cameras with four markers attached to the robot to track the IMU pose. The dataset contains the stereo images ($20 \Hz$) and IMU data ($750 \Hz$) from the Multisense S7 camera as well as the joint encoders and IMU data from the Cassie robot (at $400 \Hz$ each). 

The proposed kinematic and contact factors were implemented using the GTSAM~4.0 library~\cite{dellaert2012factor}. We utilized the built-in preintegrated IMU factor~\cite{forster2016manifold} with the iSAM2~\cite{kaess2012isam2} incremental solver. For visual odometry, we used the semi-direct visual odometry library (SVO 2.0~\cite{forster2017svo}). The Multisense camera recorded synchronized stereo images at $20 \Hz$ and IMU measurements in about $750 \Hz$. SVO processes those measurements in real-time and outputs the pose of the left camera in a fixed world frame, $\X[i]$, for the current time-step $i$. The relative transformation of the camera from time-step $i$ to $j$ can be obtained using $\Delta \X[ij] = \X[i][-1] \X[j]$. We selected keyframes approximately every $0.25$ seconds.

\begin{figure}[t] 
  \centering 
  \includegraphics[width=1.0\columnwidth]{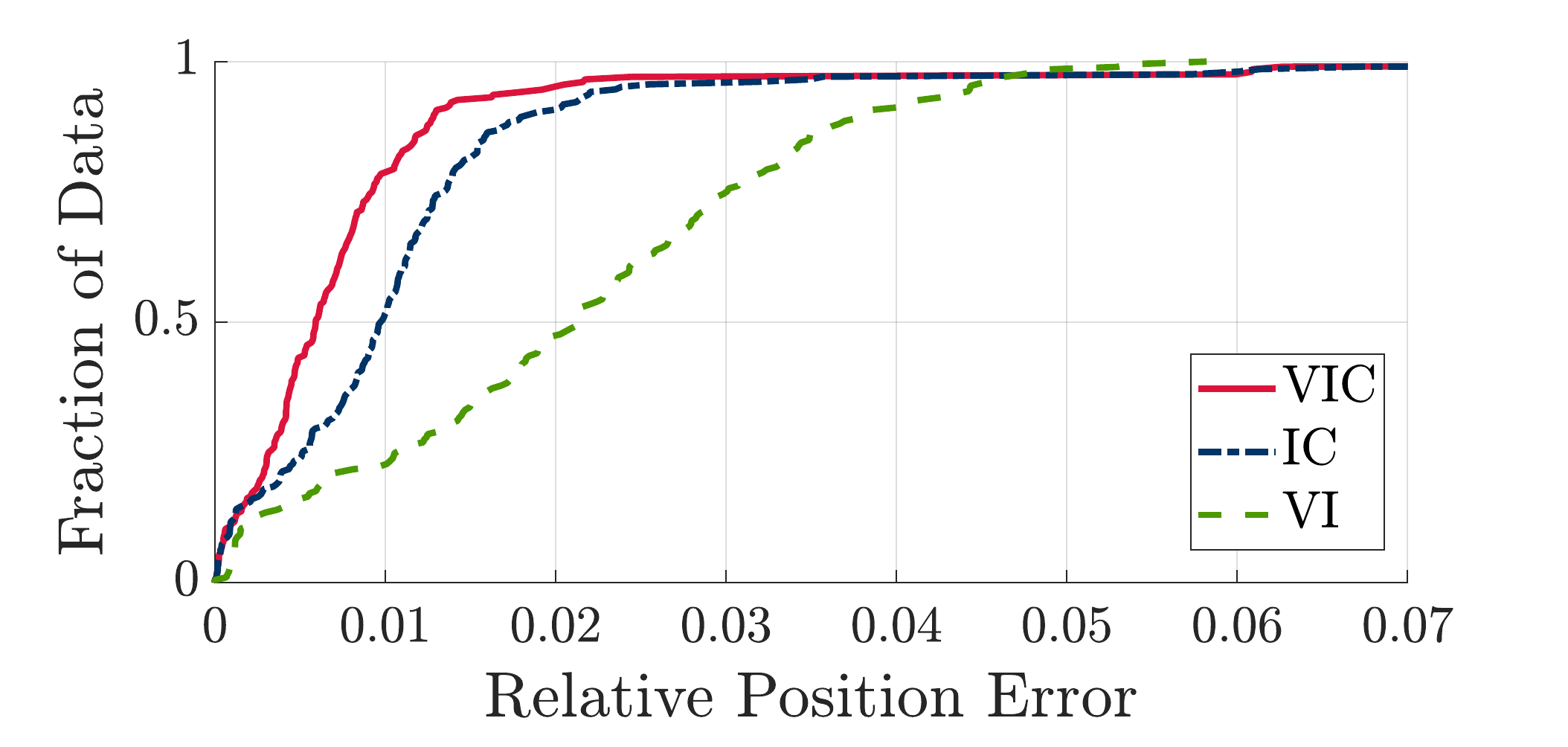}
  \caption{The Cumulative Distribution Function (CDF) of the relative position error provides a way to analyze the drift in the odometry estimates from various combinations of factors. The fraction of data corresponding to small relative position errors (low-drift odometry) is the larger for Visual-Inertial-Contact (VIC) odometry than for Inertial-Contact (IC) or Visual-Inertial (VI) odometry.  }
  \squeezeup
  \label{fig:cdf}
\end{figure}

\begin{figure}[t]
  \centering 
  \includegraphics[width=1.00\columnwidth]{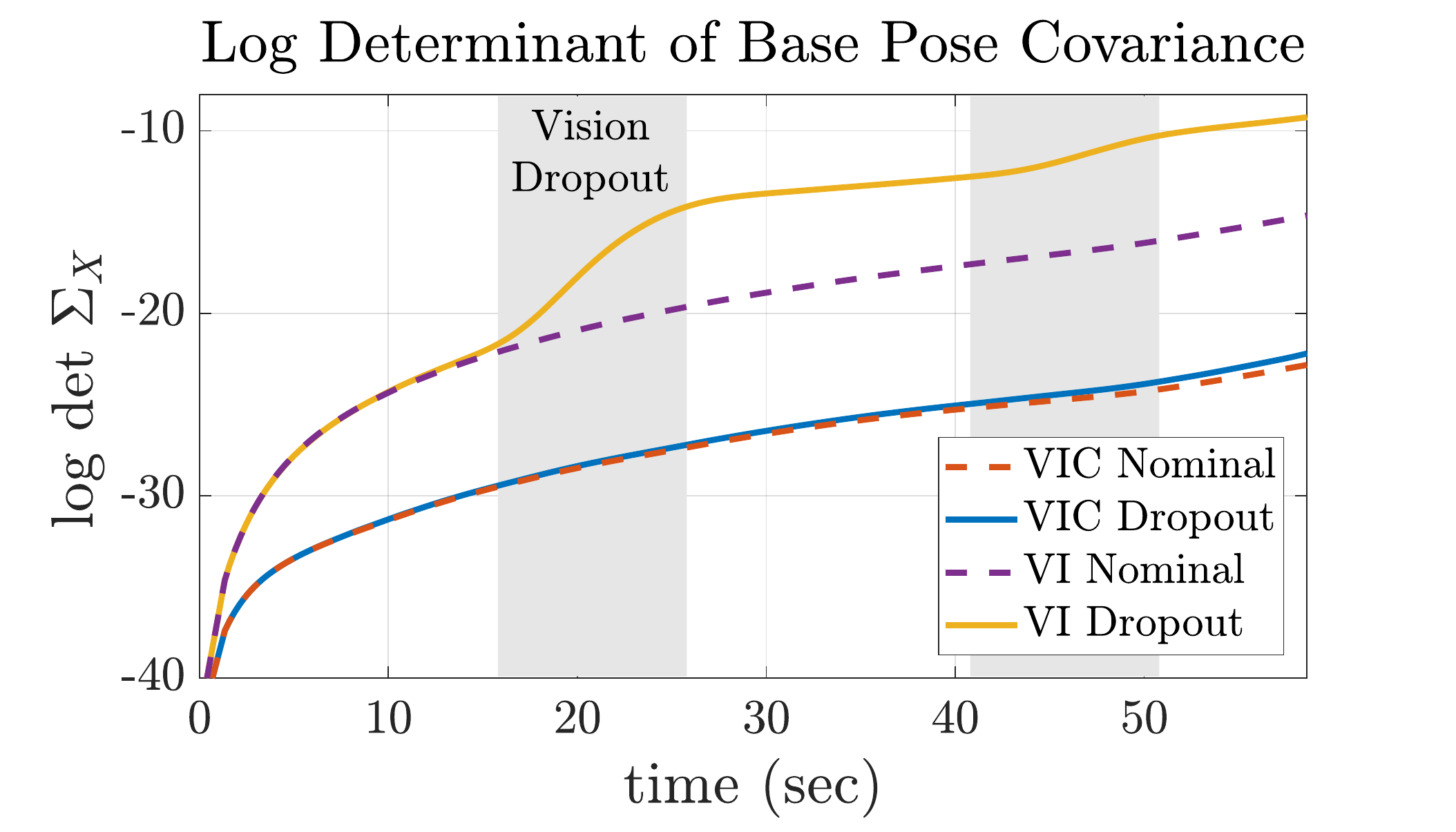}
  \caption{When vision data is lost, the covariance of the robot's base pose sharply grows for VI odometry due to the lack of additional measurements to constrain the graph. In contrast, during ``vision dropout'' periods, the additional contact factors allows the covariance estimate from VIC to remains close to the nominal case.}
  \squeezeup
  \label{fig:logdetcov}
\end{figure}

\subsection{First Experiment: Odometry Comparison}
In this experiment, we had Cassie stand in place for about 15 seconds, then slowly walk forwards and backwards along the length of the lab for approximately 45 seconds. The resulting data was used to compare the odometry performance (processed off-line) of different combinations of factors. The results are shown in Figure~\ref{fig:estimated_trajectory}. The odometry estimate from VIC, as expected, outperforms all other combinations of factors. The Cumulative Distribution Function (CDF) of the relative position error is shown in Figure~\ref{fig:cdf}. The relative position CDF provides a method for analyzing the drift of an odometry estimate.

From Figure~\ref{fig:cdf}, it can be seen that VIC has the highest fraction of data corresponding to smaller relative position errors. This means VIC has lowest drift among all  odometry systems. When the robot is walking, the hard impacts cause significant camera shake which leads to motion blur in the images. In addition, the lab environment seemingly lacked numerous quality features. These effects help to explain the relatively poor VI odometry performance. 

\subsection{Second Experiment: Vision Dropout}
One of the main benefits from including the forward kinematic and contact factors is that the state estimator can be more robust to failure of the vision system. In this experiment, we simulate the effects of ``vision dropout'' by simply ignoring SVO visual odometry data for two 10-second periods of the experimental data described in the previous section. In other words, during a ``vision dropout'' period, VIC odometry reduces to IC odometry, and VI odometry reduces down to inertial (I) odometry. Figure \ref{fig:logdetcov} shows the log determinant of the base pose covariance for VIC and VI for these ``vision dropout'' experiments. The larger the log determinant, the more uncertain the estimator is about the robot's base pose. During the ``vision dropout'' periods, uncertainty grows for VI odometry. This sharp covariance growth is due to the lack of additional sensor measurements to add into the factor graph. In contrast, the covariance growth for VIC is hardly affected over the same dropout periods. 

\subsection{Third Experiment: Terrain Factors for Loop-Closure}
Another benefit of adding the proposed contact factors comes from the addition of the contact frame poses into the robot's state. With these new state variables, it becomes simple to place additional constraints that relate the contact pose to a prior map. We test this idea on the collected experiment data by recognizing that the ground was relatively flat in the laboratory. This ``zero-height'' elevation data serves as our prior map. Figure \ref{fig:terrain} shows how adding this trivial constraint can reduce position drift in the z-direction. This experiment simply serves to illustrate the potential for ``terrain factors'', as the state estimate could be further improved if the robot was actually mapping out the terrain; there was actually a slight slope in the lab (as shown in the motion capture data). 

\begin{figure}[t]
  \vspace{2mm}
  \centering 
  \includegraphics[width=1.00\columnwidth,trim={1.0cm 0cm 1.5cm 0.5cm},clip]{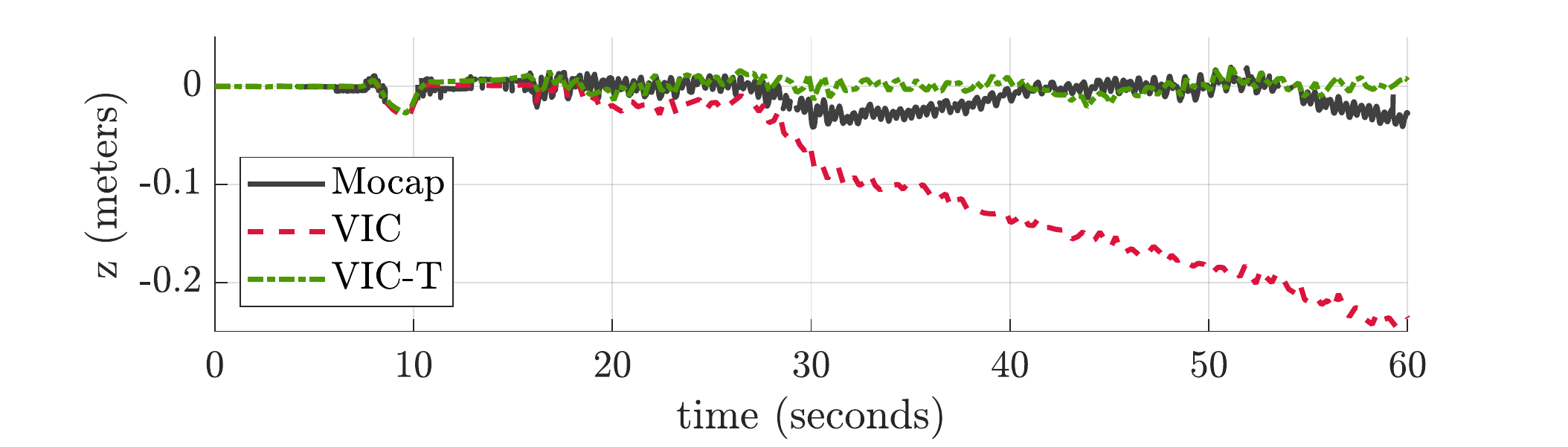}
  \caption{Since the contact pose is now part of the estimated state, it is possible to add ``terrain factors'' that relate this contact pose to a prior map. Adding the simple constraint that the contact frame z-translation is zero (VIC-T) improves the drift in the z-direction when compared to the nominal Visual-Inertial-Contact (VIC) case.}
  \squeezeup
  \label{fig:terrain}
\end{figure}

% \vspace{3.mm}
\section{Conclusion}
\label{sec:conclusion}
We developed a novel method for preintegrating contact information through an arbitrary number of contact switches using a hybrid preintegrated contact factor. The proposed approach was verified through experimental evaluations using a Cassie-series robot where a motion capture system is used as a proxy for ground truth data. Our results indicate that the fusion of contact information with IMU and vision data provide a reliable odometry system for legged robot. Furthermore, we showed that the developed visual-inertial-contact odometry system is robust to occasional vision system failures. 

In the future, we plan to incorporate loop closure constraints into our factor graph framework to further improve state estimation, paving the way for long-term mapping on legged robots. We also plan to further investigate the potential utility of ``terrain factors'' to allow the robot's state to be corrected though detected contact on an estimated map.
% \balance

%%%%%%%%%%%%%%%%%%%%%%%%%%%
% \vspace{3.mm}
\section*{ACKNOWLEDGMENT}
\footnotesize{
The authors thank Agility Robotics for designing the robot, and Yukai Gong for developing the feedback controller. We also thank Dr. Elliott Rouse and Catherine Kinnaird for providing us with the lab space and equipment necessary to collect and analyze the motion capture data. Funding for R. Hartley, M. Ghaffari Jadidi, L. Gan, and J. Huang is given by the Toyota Research Institute (TRI), partly under award number N021515, however this article solely reflects the opinions and conclusions of its authors and not TRI or any other Toyota entity. Funding for J. Grizzle was in part provided by TRI and in part by NSF Award No.~1525006.}
% \vspace{-2.5mm}

% \vspace{3.1mm}
\bibliographystyle{bib/IEEEtran}
\bibliography{bib/references}

\end{document}